\documentclass{article}
\usepackage{ijcai21}

\usepackage{times}
\usepackage{soul}
\usepackage{url}
\usepackage[hidelinks]{hyperref}
\usepackage[utf8]{inputenc}
\usepackage[small]{caption}
\usepackage{graphicx}
\usepackage{amsmath}
\usepackage{amsthm}
\usepackage{booktabs}
\usepackage{algorithm}
\usepackage{algorithmic}
\usepackage{xspace}
\usepackage{booktabs}
\usepackage[misc]{ifsym} 

\newfloat{algorithm}{tbp}{loa}
\providecommand{\algorithmname}{Algorithm}
\floatname{algorithm}{\protect\algorithmname}
\usepackage{times}
\usepackage{epsfig}
\usepackage{comment}
\usepackage{enumitem}

\newcommand{\ie}{i.\@\,e.\@\xspace}

\title{GAKP: GRU Association and Kalman Prediction for Multiple Object Tracking}

\author{
Zhen Li$^{1,2}$
\and
Sunzeng Cai$^1$\and
Xiaoyi Wang$^1$\and
Zhe Liu$^2$\And
Nian Xue$^{3,(\textrm{\Letter})}$
\affiliations
$^1$Shanghai Grandhonor Information Technology Co.Ltd\\
$^2$Nanjing University of Aeronautics and Astronautic\\
$^3$New York University
\emails
lizh0019@gmail.com,
caisunzeng@163.com,
zhe.liu@nuaa.edu.cn,
nian.xue@nyu.edu
}

\begin{document}
\maketitle
\begin{abstract}
Multiple Object Tracking (MOT) has been a useful yet challenging task
in many real-world applications such as video surveillance, intelligent
retail, and smart city. The challenge is how to model long-term temporal
dependencies in an efficient manner. Some recent works employ Recurrent
Neural Networks (RNN) to obtain good performance, which, however,
requires a large amount of training data. In this paper, we proposed
a novel tracking method that integrates the auto-tuning Kalman method
for prediction and the Gated Recurrent Unit (GRU), and achieves a
near-optimum with a small amount of training data. Experimental results
show that our new algorithm can achieve competitive performance on
the challenging MOT benchmark, and faster and more robust than the
state-of-the-art RNN-based online MOT algorithms. 
\end{abstract}

\section{Introduction}

Over the last few years, MOT (Multiple Object Tracking) technology has been playing an increasingly important role in Computer Vision (CV), which aims to extract all objects of interest automatically and obtains the corresponding motion trajectory through the spatial, temporal or visual features of video data. Although MOT is suitable to deal with complex scenes with plenty of targets, and has tremendous potential in visual monitoring/surveillance, behavior analysis, self-driving and navigation, nevertheless, it is still far behind satisfactory in complex scenarios which contains a lot of mutual occlusions and interactions of moving targets. 

\begin{figure}[t]
\centering\includegraphics[width=\columnwidth]{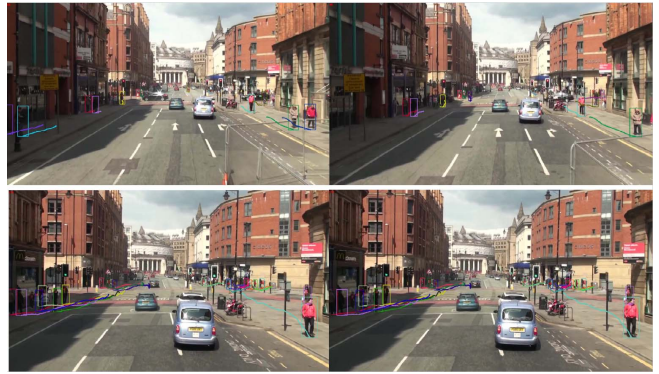} \caption{Examples of multi-object tracking results. The curved line behind
each person is the trajectory. \label{fig:illustration}}
\end{figure}

There has been a great deal of interest in designing new methods for MOT recently. The early stage of object tracking focused on single object tracking based
on feature engineering and classification, in separate steps using
conventional CV techniques. 
As an extension of single object
visual tracking \cite{Bhat2018Unveiling}, multiple object tracking \cite{Sadeghian2017Tracking} emerges as a hot issue due
to its broader practical application in complex scenarios of intelligent
surveillance recently. Typical MOT results are shown in Fig.  \ref{fig:illustration}.

The most common framework used in MOT is tracking-by-detection strategy
which links detections across frames by data association algorithms.
Under this framework, the true positions of objects in each frame
are estimated using the detector, followed by estimation of the trajectories
of multiple objects which will dynamically regenerate and disappear
depending on the detection results for different frames. 
With the rapid rise of deep learning technology, MOT has entered a new milestone \cite{Sadeghian2017Tracking,Samuel2017Deep,Zhu2018Online}.
These deep learning based approaches have improved
MOT accuracy by a large margin. However, they typically need a large
amount of training data to obtain a reasonable performance. 

For data association, either in traditional methods or deep learning
methods, most existing works realize data association by motion model
\cite{Huang2008Robust,Milan2013Continuous} or appearance model 
\cite{Yu2016POI,Wang2014Tracklet} alone. This problem has not been well studied before, partially because of the complication and variations of the features. The characteristics of the detected objects vary a great deal from different scenes, thus it is hard to robustly associate detections and predictions of an object, especially with light change, scale variation, and occlusion. 

In this paper, we propose a novel MOT method with GRU (Gated Recurrent Unit) based data association
in the framework of auto-tuning Kalman prediction, termed GAKP. To the best
of our knowledge, this paper is the first exploration to realize association
between prediction and detections using implicit motion and appearance
features, \ie, the association is done by GRU network in an end-to-end
manner, without explicitly weighting factors of motion and appearance
features as in \cite{Yu2016POI,Sadeghian2017Tracking}. Experimental
results show that our implicit data association outperforms the state-of-the-art
explicit data association, while not introducing extra computational cost.

Our primary contribution is manifold:
\begin{itemize}
\item We integrate GRU for data association in the framework of the auto-tuning
Kalman prediction to take advantage of deep learning and compensate
the disadvantage: the Kalman tracker is efficient while data association
accuracy is improved by GRU based on numerous online training data. 
\item We utilize GRU to achieve an accurate and robust association between predictions
and detections, by using various features including motion feature,
spatial-feature, deep feature and so on. The mapping from various
features to the association similarity is optimized by GRU in an end-to-end
manner. 
\end{itemize}

\begin{figure}[t]
\centering\includegraphics[scale=0.36]{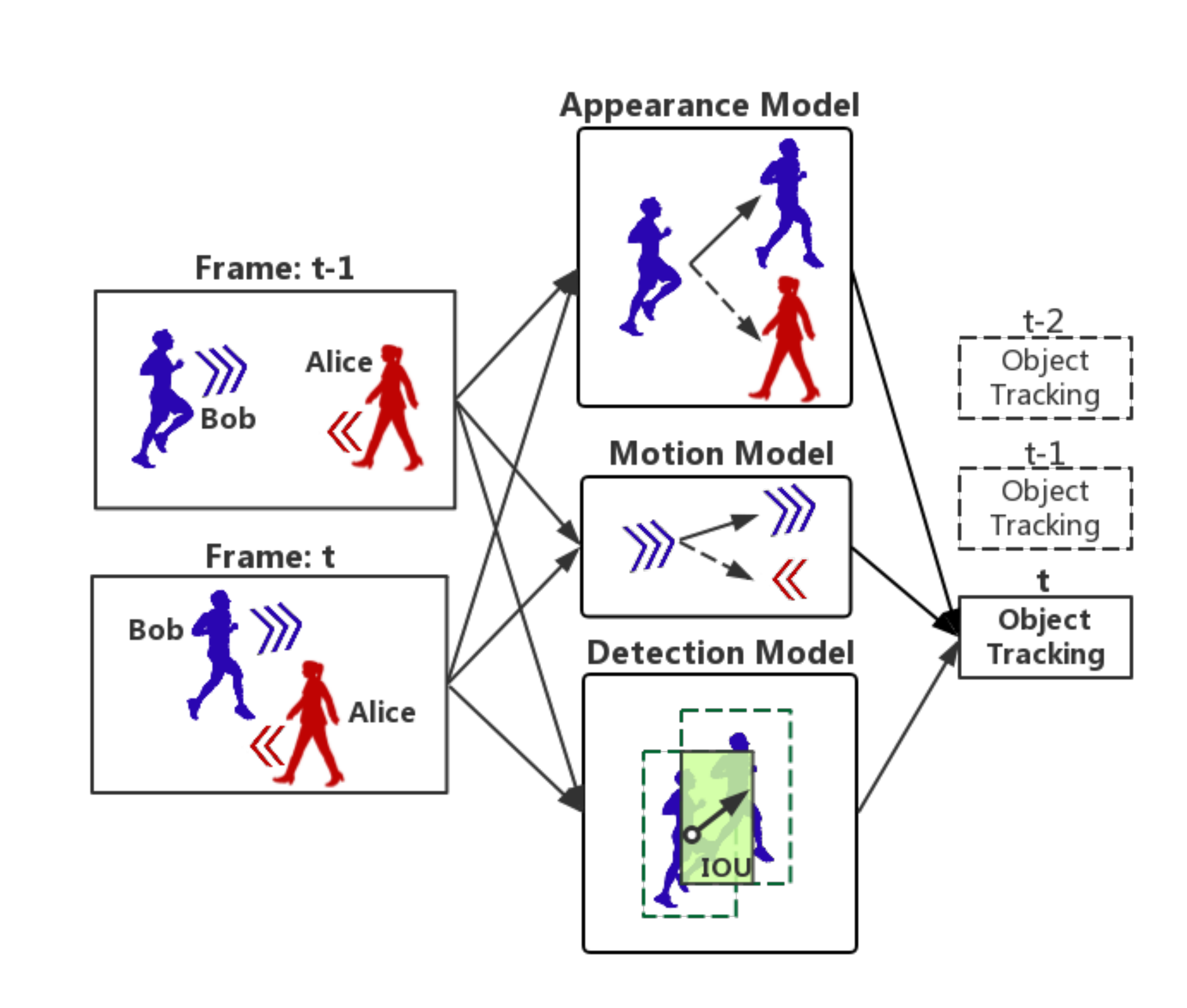} \caption{The proposed MOT utilizes motion model, appearance model, as well as
detection bounding box information. \label{fig:MOT-GRU-2}}
\end{figure}

\section{Related Work}

MOT methods can be categorized into online and offline modes according
to different application requirements. Thereinto, offline MOT algorithms
can access the entire frames of video and utilize both past and future
frames to optimize trajectories. Therefore, it can be regarded as
an optimization problem to find a set of trajectories with the minimum
global cost function, which can be solved by standard Linear Programming
techniques in \cite{Berclaz2009Multiple}.
Common offline detection
association can be formulated as a Maximum A Posteriori (MAP) problem
and solved by the Hungarian algorithm 
\cite{bewley2016simple}.
In general, offline tracking can achieve higher tracking accuracy
compared with online methods, at the cost of more computational complexity.
In contrast, the online MOT methods are desired in real-time scenarios,
as they merely exploit the information available no later than the
current frame. 

\begin{figure*}[t]
\centering \includegraphics[width=1.95\columnwidth]{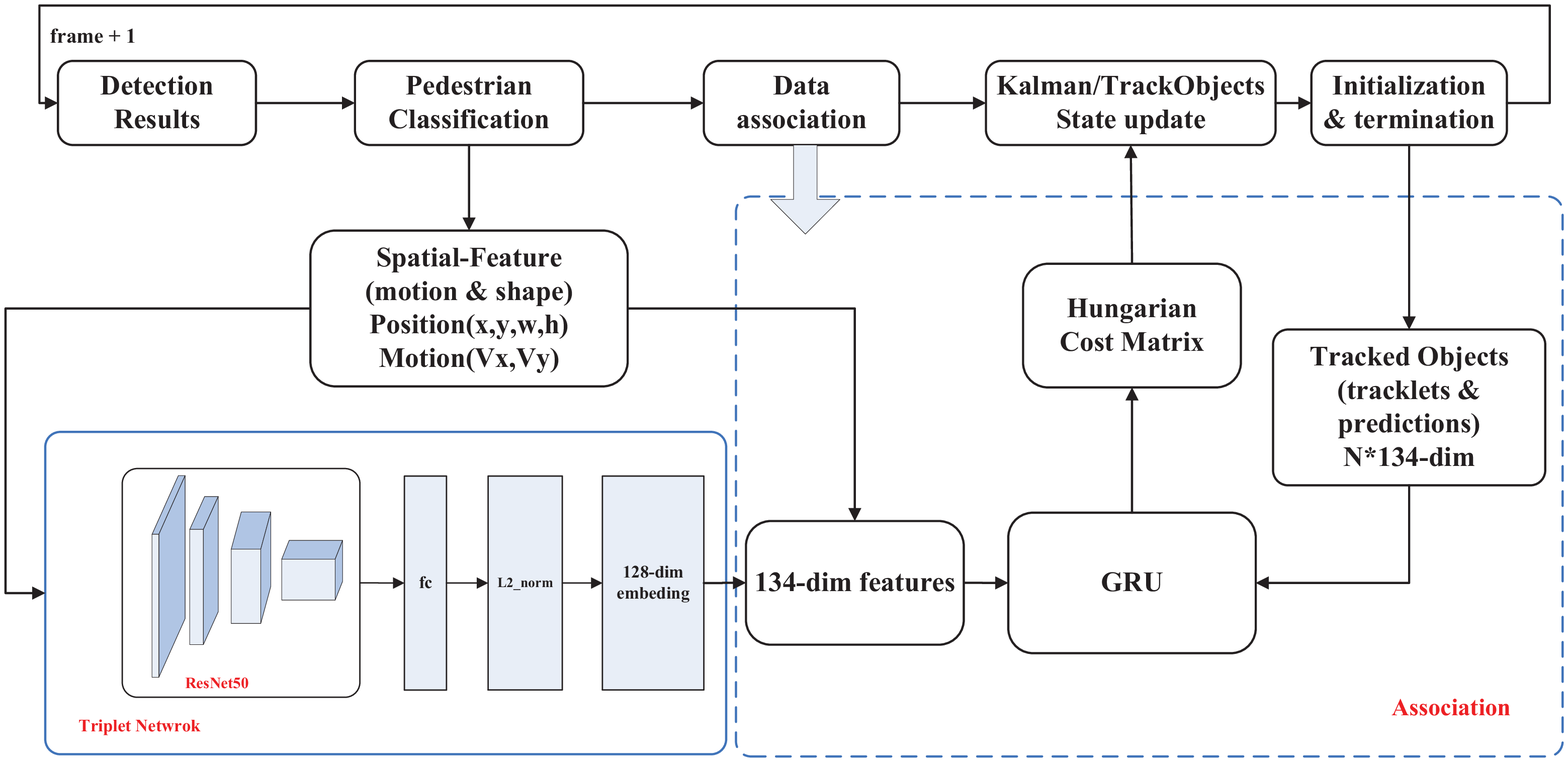}
\caption{The flow chart of the proposed MOT framework. Our online MOT algorithm
consists of three parts as follow: 1) Gaussian statistics (mean and covariance) are given
by auto-tuning Kalman prediction; 2) data association is realized
by using Hungarian algorithm; and 3) The similarity between detections
and predictions is evaluated through the recurrent neural network
with GRU, which is trained with the ground-truth of the MOT$17$ data \protect\cite{Milan2016MOT16}. \label{fig:framework}}
\end{figure*}

This work falls into the category of online MOT, and we focus on improving
the data association between detections and predictions. The key issue
of association is how to obtain correct associations robustly with
feature variations. Existing works realize data association mainly
by three types of models: motion model alone, appearance model alone,
and the combination.

\textbf{Motion Model.} The motion model describes how a target moves.
The key of this model is that a more precise prediction of targets in the future frames will reduce the search space of the association
model and thus increase the matching accuracy. Popular motion models include
linear and non-linear motion models. Linear motion models follow a
linear movement with constant velocity across frames, which is the
early stage popular models in MOT \cite{Breitenstein2009Robust}.
Non-linear motion models are proposed to produce a more accurate prediction \cite{Dicle2013The}.
In recent years, the depth recurrent neural network (RNN) method is a trend for
a non-linear approach for MOT motion prediction. However, as a common problem
of using RNN implementation, a large amount of training data is required
for optimal performance. In the meanwhile, for complex scenes, the
amount of training trajectory data is far from enough, which may result
in over-fitting.

\textbf{Appearance Model.} 
In early years, some approaches use color histogram, covariance matrix representation, pixel comparison representation, SIFT-like features, or pose features \cite{Choi2010Multiple,hong2014Visual,Izadinia2013Multi}.
Deep learning based models have emerged as a very powerful tool to
deal with different kinds of vision challenge including image detection
and classification. The strong observation model provided by the deep
learning model for target detection can boost the tracking performance
significantly \cite{Yu2016POI,Lee2016Multi}. Deep neural network
architectures have been used for modeling appearance recently. In
these architectures, high-level features are extracted by convolutional
neural networks trained for a specific task and achieve a significant
improvement.

\textbf{Composite Model.} Some recent works attempt to combine the
motion model and appearance model together to enhance the association
accuracy. A composite model of hand-crafted feature with position,
size and appearance feature is defined in \cite{Yu2016POI}, which
provides a competitive performance. However, hand-crafted feature
has a disadvantage that it is difficult to tune the weights of each
component to be robust in different scenarios. For example, the tracker
using only IoU (Intersection-over-Union) is not effective for high-speed
small target tracking, as the IoU between the target and the detection
easily reaches zero, while the tracking using only Euclidean distance
is not reliable for large targets due to the error and deformation
of the tracklets. Thus a combination of appearance feature is
a reasonable direction to improve the robustness of data association.
Despite extensive experimentation with RNN-LSTM architectures in
\cite{Sadeghian2017Tracking}, the learned metric did not perform
as well as the simpler hand-crafted functions, presumably due to the
small size of the training set. 

In our work, we integrate the auto-tuning
Kalman method for prediction step and GRU for the association step.
The link probability between predictions and detections is predicted
with non-linear combination of motion and appearance features. This method significantly improves the tracking performance while reducing the computational
cost. The feature models utilized in the proposed GAKP framework
is illustrated in Fig.  \ref{fig:MOT-GRU-2}.




\begin{figure*}[t]
\centering \includegraphics[width=1.75\columnwidth]{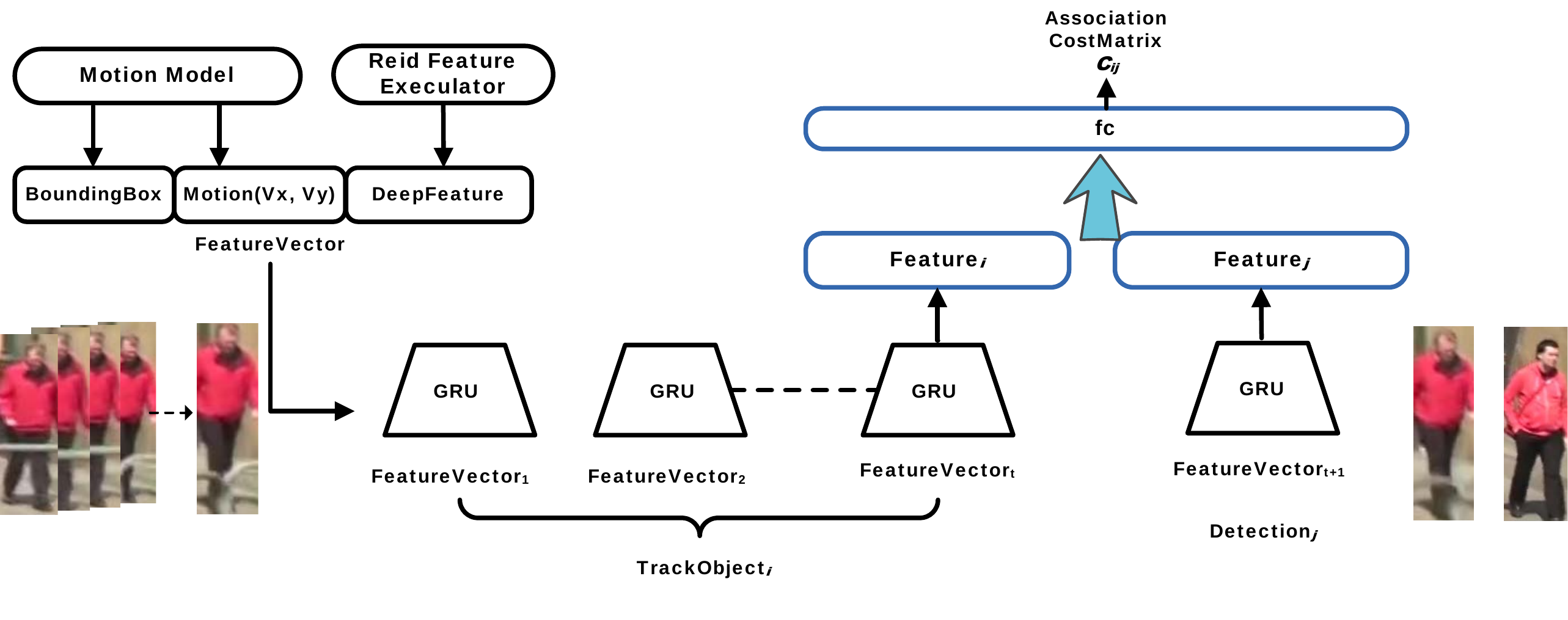}
\caption{Flow chart of the Cost Matrix calculation with GRU. 1) The feature vector
comes from the motion model and feature extractor. 2) Feature vector
of tracked object $i$ in frame $t$ and the feature of detection
$j$ in frame $t+1$ are used as input of the GRU model to calculate
the association cost $C_{ij}$ of the cost matrix.\label{fig:framework_gru}}
\end{figure*}

\section{Online MOT Algorithm}

In this section, we describe our proposed MOT tracker with GRU data-association and auto-tuning
Kalman prediction method. 
The flow chart of the proposed GAKP is depicted, including the motion model (sec. \ref{subsec:Motion-Model}), appearance model (sec. \ref{subsec:Appearance-Model})
and the end-to-end data association (sec. \ref{subsec:Data-Association})
which will be elaborated in following subsections. Finally, the proposed
GAKP algorithm is summarized.

\subsection{Proposed Framework \label{subsec:Proposed-Framework}}

The overall GAKP framework is shown in Fig.  \ref{fig:framework},
where spatial feature and motion feature are obtained from detection
and Kalman filter respectively, and the appearance feature is extracted
by Triplet ResNet-$50$ Network in \cite{Hermans2017In}. The end-to-end
data association module is shown in Fig.  \ref{fig:framework_gru}.
The key components of the multi-object tracker are listed as follows: 

\textbf {Detection and bounding box processing.} The pedestrian detection
responses are processed by the classification to select high-quality
pedestrian bounding boxes. 

\textbf {Motion Prediction.} Based on the previous tracked object at frame
$t-1$, we predict the likely location of each target at frame $t$
via the motion model, and use the detection results to initialize
tracklets at frame 0. We use auto-tuning Kalman prediction since it can
achieve good performance in near-linear motion system with small amount
of training data instead of RNN. 

\textbf {Appearance Features.} The cropped images of pedestrians are fed
into feature extractor to get the $128$-dim appearance feature embeddings.
Meanwhile, it should be noticed that not only the deep feature but
also all the other features can be used as inputs to calculate the
data association probability in our framework, such as color feature
and position keypoints. 

\textbf {Association Cost Matrix.} As the core step of our algorithm,
we propose to learn a GRU model to estimate
the association cost functions. The training data are derived from
the ground-truth of MOT$17$ challenge \cite{Milan2016MOT16}. The
link probability is calculated from the features of predictions and
detections. 

\textbf {Data Association.} Given the end-to-end estimated association
cost matrix, we perform data association via a simple yet effective
Hungarian algorithm. 

\textbf {State Update.} The matching result and cost function items $C_{ij}$
will be fed to auto-tuning Kalman module to update the object motion
states. 

\subsection{Motion Model \label{subsec:Motion-Model}}

It is well known that Kalman filter is an effective approach \cite{Kalman1961New,Bar2001Estimation}
to find the optimal estimation of near-linear motion states. The predicted
mean and covariance states are given by 
\begin{equation}
\hat{x}_{k/k-1}'=\mathbf{F}_{k}\hat{x}_{k-1/k-1},\label{mean_prediction}
\end{equation}

\begin{equation}
\hat{\mathbf{P}}_{k/k-1}'=\mathbf{F}_{k}\hat{\mathbf{P}}_{k-1/k-1}\mathbf{F}_{k}^{T}+\mathbf{Q}_{k}^{*},\label{covariance_prediction}
\end{equation}
where $k$ indicates the index in time series, $\hat{x}_{k/k-1}'$
is predicted object statement, $\hat{\mathbf{P}}_{k/k-1}'$ is predicted
object covariance, $\mathbf{F}_{k}$ is state-transform matrix, $\mathbf{Q}_{k}^{*}$
is system prediction error.

The motion mean and covariance update is given by 
\begin{equation}
\mathbf{K}_{k}=\frac{\mathbf{P}_{k/k-1}\mathbf{H}_{k}^{T}}{\mathbf{H}_{k}\mathbf{P}_{k-1/k-1}\mathbf{H}_{k}^{T}+\mathbf{R}_{k}^{*}}\label{K_gain}
\end{equation}

\begin{equation}
\hat{x}_{k/k}=\hat{x}_{k/k-1}'+\mathbf{K}_{k}(\mathbf{Z}_{k}-\mathbf{H}_{k}\hat{x}_{k/k-1}')\label{mean_update}
\end{equation}

\begin{equation}
\hat{\mathbf{P}}_{k/k}=(\mathbf{I}-\mathbf{K}_{k}\mathbf{H}_{k})\hat{\mathbf{P}}_{k/k-1}(\mathbf{I}-\mathbf{K}_{k}\mathbf{H}_{k})^{T}+\mathbf{K}_{k}\mathbf{R}_{k}^{*}\mathbf{K}_{k}^{T}\label{covariance_update}
\end{equation}
where $\mathbf{K}_{k}$ is the Kalman gain which can balance the prediction
and detection to cancel the noise, resulting in filtered states $\hat{x}_{k/k}$
and $\hat{\mathbf{P}}_{k/k}$. According to the formula of Kalman
gain $\mathbf{K}_{k}$, the optimal motion state information $\hat{x}_{k/k}$
of the target at the current time is obtained. $\mathbf{H}_{k}$ is
the transfer matrix between target motion state and measured position
state.
When compared with the target
position state, both the predicted
target state $\hat{x}_{k/k-1}'$ and the detected target state $\mathbf{Z}_{k}$
are expected to have errors, \ie, the system prediction error $\mathbf{Q}_{k}^{*}$
and the target detection error $\mathbf{R}_{k}^{*}$, respectively.

Both $\mathbf{Q}_{k}^{*}$ and $\mathbf{R}_{k}^{*}$ are deterministic
noise matrices, and Kalman filter automatically guarantees statistical
consistency when the full structure of the system state $\left(\mathbf{F}_{k},\mathbf{H}_{k},\mathbf{Q}_{k}^{*},\mathbf{R}_{k}^{*}\right)$
is known. However, in many situations the model is not known precisely
and the Kalman filter must be tuned. It is hard to tune the coefficients
of Kalman filter (such as process noise $\mathbf{Q}_{k}^{*}$, and
measure noise $\mathbf{R}_{k}^{*}$ which defined in Eq. \ref{covariance_prediction}
and Eq. \ref{covariance_update}), e.g. significant effort is required
to tune various Kalman filter models for non-white noise. An auto-tuning
$\left(\mathbf{Q}_{k},\mathbf{R}_{k}\right)$ with Bayesian Optimization
is proposed to minimize normalized estimation error squared (NEES)
in \cite{AutoKalman2018}. However, in \cite{AutoKalman2018} only
the motion characteristics are considered to estimate the errors.
In this work, we demonstrate that, by considering both the motion
characteristics and visual similarity, a more accurate estimation
of system prediction error and target detection error can be obtained,
resulting in a better performance of Kalman tracking. Specifically,
we propose a new version of the system prediction error $\mathbf{Q}_{k}^{*}$
and the target detection error $\mathbf{R}_{k}^{*}$ as follow:

\begin{equation}
\mathbf{Q}_{k}^{*}=\frac{\mathbf{Q}_{k}}{C+\lambda_{c}}\label{eq:Q*}
\end{equation}

\begin{equation}
\mathbf{R}_{k}^{*}=\frac{\mathbf{R}_{k}}{C+\lambda_{c}}\label{eq:R*},
\end{equation}
where $\mathbf{Q}_{k}$ and $\mathbf{R}_{k}$ are the estimated errors
given by \cite{AutoKalman2018} which considers motion characteristics,
and $C\in\left(0,1\right)$ is the composite similarity (link probability)
between predictions and detections given by GRU deep learning, where
a high $C$ value indicates that the measured detection is more reliable
in Kalman gain update progress, and $\lambda_{c}$ is a small factor
for regularization.

Mahalanobis distance \cite{Wojke2017simple} is utilized in this work
to improve the Euclidean distance between predicted Kalman states
and detected measurements, which is defined as follow: 
\begin{eqnarray}
D\left(i,j\right)=(\mathbf{Z}_{k}\left(j\right)-\hat{x}_{k/k}\left(i\right))^{T}\mathbf{S}_{i}^{-1}(\mathbf{Z}_{k}\left(j\right)-\hat{x}_{k/k}\left(i\right))\label{Mahalanobis}
\end{eqnarray}
where $\mathbf{S}_{i}=\mathbf{H}_{k}\mathbf{P}_{k/k}\left(i\right)\mathbf{H}_{k}^{T}$.
The measurement space of the $i$-th track is denoted by Multivariate
Gaussian Distribution $\left(\hat{x}_{k/k}\left(i\right),\mathbf{S}_{i}\right)$.
We keep the candidates where the Mahalanobis distances are within
$95\%$ confidence interval computed from the inverse noise $\chi^{2}$
distribution, and the threshold is $9.4877$ for 4-dimensional Mahalanobis
distance. Hungarian algorithm is then used to match pairs after the
pre-filtering by Mahalanobis distances.

\subsection{Appearance Model \label{subsec:Appearance-Model}}

The underlying idea of the appearance model is that the similarity
score can be computed between a target and candidate detection based
on visual features. 
Re-identification networks \cite{Chen2017Beyond,Schroff2015FaceNet,Hermans2017In}
can be utilized by learning a similarity metric so that the target of
the same identity is closer to each other than different identities in
the embedded feature space. 
The appearance feature extractor of our model is ResNet-$50$ which is a pre-trained person re-identification model proposed in \cite{Hermans2017In}. It is robust to occlusions
and other visual disturbances. The triplet loss for training the CNN
is defined as follow: 
\begin{equation}
\mathcal{L}_{trip}=\sum_{\left(z_{a},z_{p},z_{n}\right)\in Z}\max\left(0,d_{a}\left(z_{a},z_{p}\right)-d_{p}\left(z_{a},z_{n}\right)+\theta\right),
\end{equation}
where $\left(z_{a},z_{p},z_{n}\right)$ denotes an instance of triplet
where $z_{a}$ is the anchor, $z_{p}$ is a candidate of positive
samples, and $z_{n}$ is a candidate of negative examples, $d_{a}\left(z_{1},z_{2}\right)$
denotes the euclidean distance between $z_{1}$ and $z_{2}$ called
the appearance distance. The convolutional feature maps of original
target images are flattened, fed into the fully connected layers and
finally normalized by an l2-normalization layer. The output is the
$1024$-dimensional appearance embedding $z$.

\subsection{Data Association \label{subsec:Data-Association}}

In MOT framework, data association is an important part to define
the correspondence between detections and tracking hypotheses object
on the basis of the predicted motion state and visual features. A
baseline tracklet association framework is presented in \cite{Huang2008Robust},
and many improved algorithms followed this framework. With the independence
assumption, the object association can be formulated as follow: 
\begin{eqnarray}
\mathcal{S}^{*}=\arg\max_{\mathcal{S}}\prod_{T_{i}^{L}\in\mathcal{T}^{\mathcal{L}}}P\left(T_{i}^{L}/\mathcal{S}\right)\prod_{S_{k}\in\mathcal{S}}P\left(S_{k}\right),
\end{eqnarray}
where $S_{k}=\left\{ T_{i0}^{L},T_{i1}^{L},...,T_{il_{k}}^{L}\right\} $
is a set of tracklets, $l_{k}$ is the number of tracklets in $S_{k}$,
and $\mathcal{S}=\left\{ S_{k}\right\} $ is the tracklet association
set. A conventional link probability between tracking object and detection
is proposed in \cite{Huang2008Robust}: 
\begin{eqnarray}
P_{link}\left(r_{j}|r_{i}\right)=A_{pos}\left(r_{j}|r_{i}\right)A_{size}\left(r_{j}|r_{i}\right)A_{appr}\left(r_{j}|r_{i}\right),
\end{eqnarray}
where $A_{pos}\left(r_{j}|r_{i}\right)$, $A_{size}\left(r_{j}|r_{i}\right)$,
$A_{appr}\left(r_{j}|r_{i}\right)$ is position, size and appearance
link probability between tracking object $i$ and detection object
$j$. The cost function calculated with the link probability and the optimal solution of cost matrix are respectively as follow,
\begin{eqnarray}
C_{ij}=\textrm{ln}P_{link}\left(r_{j}|r_{i}\right),
\end{eqnarray}
\begin{eqnarray}
j^{*}=\arg\min_{j}\frac{1}{\mathcal{T}}\sum_{i}^\mathcal{T}C_{ij}.\label{eq:optimal_cost}
\end{eqnarray}


\subsubsection*{Explicit Feature Association \label{cost_explicit}}

Explicit features refers to the commonly used features such as motion
features (obtained from Kalman filter), bounding boxes, and visual
content features. When the explicit features are combined properly
by RNN method, they can effectively improve the accuracy and robustness
of MOT in complex scenarios. The cost-function of prediction and detection
with different features is defined as follow:

\begin{equation}
C_{ij}=|D_{ij}|^{2}+\lambda_{\textrm{I}}|\textrm{IoU}_{ij}|^{2}+\lambda_{v}|\triangle v|^{2}+\lambda_{a}|\triangle a|^{2}+\lambda_{f}|\triangle f|^{2},\label{eq:cost_function}
\end{equation}
where $\lambda$ denotes the weight of each feature, and the explicit
features between tracking object and prediction object as follow:
$D_{ij}$ denotes the distance, $\textrm{IoU}_{ij}$ denotes intersection-over-union
($bbox_{i}\cap bbox_{j}/bbox_{i}\cup bbox_{j}$), $\triangle v=v_{i}-\hat{v}_{ij}$
and $\triangle a=a_{i}-\hat{a}_{ij}$ denotes the velocity error and
accelerate error, respectively, between predicted hypotheses and the
detections pairs, and $\triangle f=f_{i}-\hat{f}_{j}$ denotes the
deep feature distance between tracking and detection.

The optimal weights of explicit features are predicted by trained RNN in the proposal. The features are fed into RNN, and the output is the optimal weights $\lambda$ of each feature. 
For training the RNN, the datasets is generated as:
the explicit features are calculated with the pairwise of groundtruth and truth positive detection, where the truth target detections are chosen with the maximum overlap of groundtruth. 
Given $\mathcal{N}$ pairwise training datasets, the cost function with Adam descent algorithm going to be minimized is formulated as: $\min\frac{1}{\mathcal{N}}\sum_{i}^{\mathcal{N}}C_{i}$. 


\subsubsection*{End-to-end Implicit Feature Association by GRU\label{cost_GRU}}

Combining explicit features linearly is not the best way to compute
the similarity score, as these features are not independent. Instead,
we propose an end-to-end mapping from the input data to the solution
of the data association problem. The end-to-end data association module
is shown in Fig.  \ref{fig:framework_gru}. The composite features,
including spatial, motion and deep features, will be used for both
training and prediction. The combined feature vector pairs of predictions
and detections are fed into GRU, and similarity of the feature
pair is the output. Specifically, we encode long-term dependencies
in the sequence of observations by using GRU networks which is shown
as follow: 
\begin{equation}
\begin{cases}
\begin{array}{l} 
r_{t}=\sigma(W_{r}\cdot[h_{t-1},x_{t}])\\
z_{t}=\sigma(W_{z}\cdot[h_{t-1},x_{t}])\\
\tilde{h}_{t}=\textrm{tanh}(W_{h}\cdot[r_{t}\ast h_{t-1},x_{t}])\\
h_{t}=(1-z_{t})\ast h_{t-1}+z_{t}\ast\tilde{h}_{t}\\
y_{t}=\sigma(W_{o}\cdot h_{t}),
\end{array}\end{cases}\label{eq: GRU}
\end{equation}
where $z_{t}$ and $r_{t}$ denote update-gate and reset-gates. The
ground truth sequence is used to train GRU cell with an online manner
that will be described as follow.

The GRU network, as other deep learning networks, requires a large
amount of training data to obtain a reasonable performance. 
In view of this, the training data for association is acquired in an
online generation process, such that only a small amount of video
data will provide a great variety for good generalization. 
The highest scoring detections with $\textrm{IoU}>0.5$ overlap of the ground truth are
labeled as the positive samples, the maximum overlaps of those having $\textrm{IoU}<0.5$ are labeled as the negative samples. And we randomly crop the sample images with $0.8\sim1.2$ times of the target size around them to augment the training data.
The cross entropy loss function is used to train the GRU network to
predict the similarity, with a gradient descent optimization algorithm
of Adam. The output score $0\sim1$ indicates the matching similarity
between the detection result and the tracking target.


\subsection{Proposed MOT Algorithm \label{subsec:algorithm}}

The whole procedure of the proposed MOT algorithm, integrating the
auto-tuning Kalman prediction and GRU association (sec. \ref{subsec:Motion-Model}-\ref{subsec:Data-Association}),
is summarized in Algorithm \ref{GAKP}.

\begin{algorithm}[t]

\noindent \caption{Proposed GAKP Algorithm\label{GAKP} }
\textbf{Input}: 
Video frame $V = I_1, \cdots, I_T$ 

\textbf{Output}: 
The tracking trajectories $\mathcal{T}_t^{\mathcal{L}}$ in $t$-th frame; 

\textbf{Initialization}: Initialize new trajectories $\mathcal{T}_{t=1}^{\mathcal{L}}$ with detections, and set the model and appearance features.

\hfill{}

\textbf{Repeat}: For $t=1,...,T$ 
\begin{enumerate}\setlength{\itemsep}{0pt}
\item Detect boxes $\mathcal{D}^N_t$ with input image $I_t$;
\item Extract motion features $\{b_t\}_{j=1}^N$ with motion model \ref{subsec:Motion-Model};
\item Predict features of each target in next frame with Eq. \ref{mean_prediction} and Eq. \ref{covariance_prediction}; Find the gating threshold of Mahalanobis distance $D\left(i,j\right)$ with Eq. \ref{Mahalanobis};
\item Extract appearance feature $\{f^t\}_{j=1}^N$ with pre-trained ResNet-$50$ appearance model \ref{subsec:Appearance-Model};
\item for all $i \in \mathcal{T}^\mathcal{L}_{t-1}$ do:
\item[] Compute cost matrix $C_{ij}$ using GRU association model Eq. \ref{eq: GRU} for all $j \in N$;
\item Gate the cost matrix with threshold of Mahalanobis distance which calculated in motion model;
\item Associate $\mathcal{T}_{t-1}^\mathcal{L}$ with $\mathcal{D}_t^N$ using Hungarian algorithm \ref{subsec:Data-Association};
\item Initialize new trajectories with unassociated detections;
\item Update $\mathcal{T}_t^\mathcal{L}$
\end{enumerate}
\end{algorithm}

\section{Experiments}

In this section, we use our learned proposed algorithm to tackle the multi-object tracking problem. The overall performance of our framework compared with the other trackers is evaluated on the MOT challenges \cite{Milan2016MOT16}.

\begin{figure}[t]
\centering \includegraphics[width=\columnwidth]{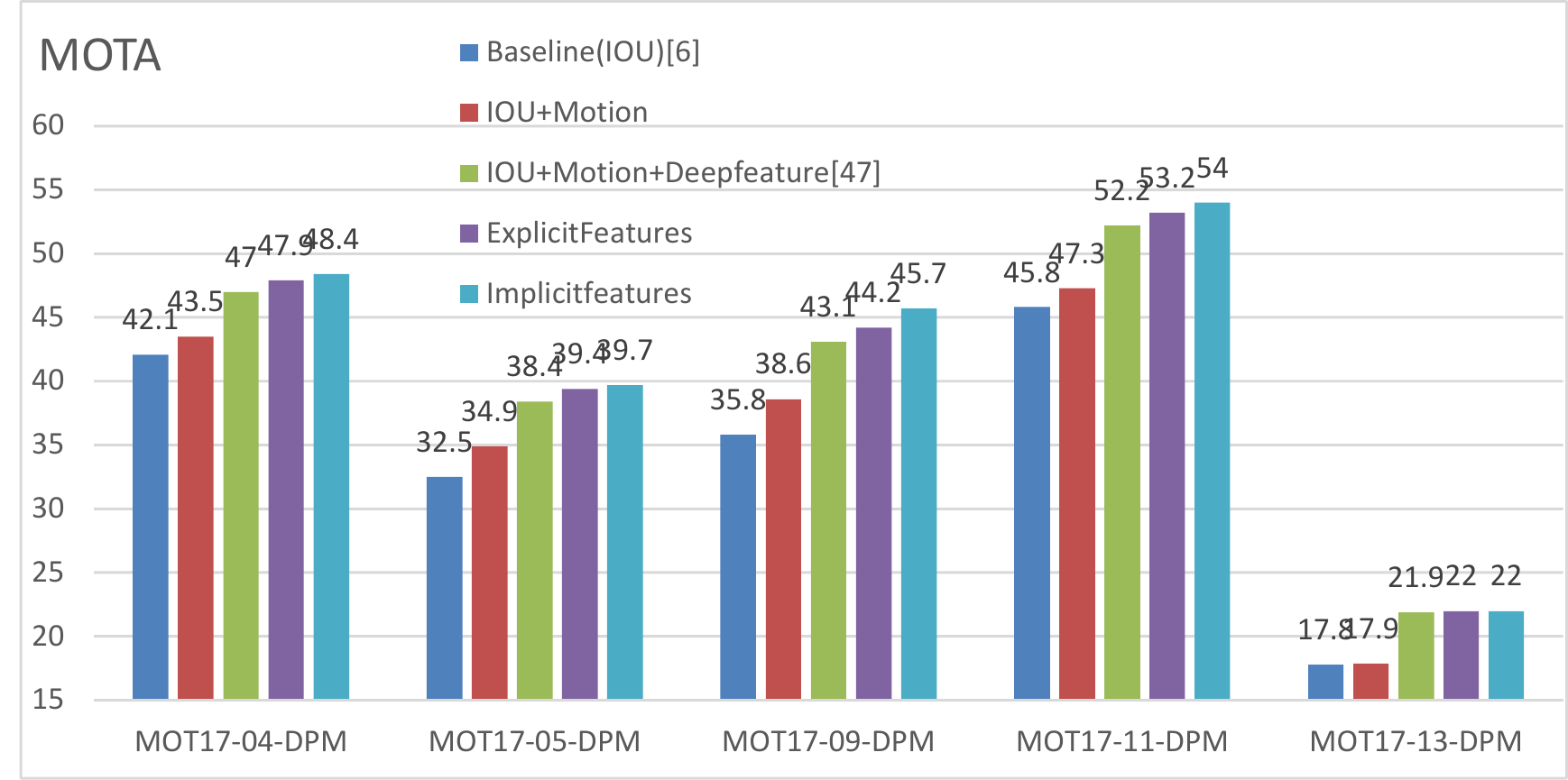}\caption{Tracking result with MOT17-DPM training data compared with baseline
approaches. 1) Baseline: IOU-based; 2) IOU and motion based association; 3) Combine the spatial feature, deep features and IOU; 4) Explicit features: the explicit weights of each
feature are learned from network; 5) Implicit features: the cost function
is obtained from the end-to-end learning GRU network.\label{fig:compare} }
\end{figure}

\subsection{Datasets and Protocols}

 The MOTchallenge benchmark includes MOT$2015$ \cite{Laura2015MOT15}, MOT$2016$ and MOT$2017$ \cite{Milan2016MOT16}.  We evaluate our approach on the MOT$16$ and MOT$17$ Benchmarks. MOT$16$ offers 
$14$ video sequences ($7$ for training and $7$ for testing) which
are captured by static and moving cameras.  MOT$17$ provides the same sequences as MOT$16$, but each sequence provides $3$ different detection results, DPM \cite{felzenszwalb2010dpm}, Faster R-CNN \cite{Ross2015Fast} and SDP \cite{Yang2016Exploit}, researchers are asked to submit tracking results with these detectors.
 
 For evaluation, the metric multi-object tracking accuracy (MOTA) provides the combination of the False Positive (FP), False Negative (FN) and ID switch (IDs) amongst all the trajectories against the Ground Truth (GT).
\begin{eqnarray}
\textrm{MOTA}=1-\frac{\textrm{FP}+\textrm{FN}+\textrm{IDs}}{\textrm{GT}}
\end{eqnarray}
 There are other metrics including Mostly Tracked (MT) and Mostly Lost (ML) that provide an indication of the trajectory fragmentation and processing speed (frames per second, FPS), respectively. 

\subsection{Implementation Details\label{Implementation}}

For appearance feature extraction, we employ the deep CNN with ResNet-$50$ backbone pre-trained using triplet loss in \cite{He2015Deep}. The cropped image is resized to $256\times128$. ReLU is used for activation
and Adam optimizer is used for the network training. The number of cell for each
GRU is $134$, which is a concatenation of $128$-dim deep feature and $6$-dim spatial-feature.
GRU network is trained with mini-batch size of $64$. Learning
rate is initialized as $0.002$, with a decay rate $0.1$ every $20$ epochs. The regularization parameter $\lambda_c$ in Eq. \ref{eq:Q*} and Eq. \ref{eq:R*} is set to 0.5 according to preliminary experiments. In all experiments, the value of parameters GRU Hidden-size $H$ and sequence length are $134$ and $7$, respectively. 

The simulation environments are given as follows: Tensorflow, Ubuntu 16.04, Intel\textsuperscript{\textregistered}
Xeon\textsuperscript{\textregistered} CPU E5-2667 v4 @ 3.20GHz \texttimes{} 32,
64GB RAM, and NVIDIA\textsuperscript{\textregistered} GeForce\textsuperscript{\textregistered}
GTX 1080 Ti/PCIe/SSE2.

\begin{table*}[tbh]
\caption{Tracking result with MOT17-DPM training data compared with baseline
approaches\label{tab:MOT_result-self} }
\centering{} %
\begin{tabular}{|c|c|c|c|c|c|c|c|}
\hline 
Trackers  & MOTA$\uparrow$  & ID F1 $\uparrow$  & MT $\uparrow$  & ML $\downarrow$  & FP$\downarrow$  & FN$\downarrow$  & IDs$\downarrow$ \tabularnewline
\hline 
Baseline(IOU) \cite{Bochinski2017High}& 35.1 & 37.2 & 49 & 255 & 5901 & 66031 & 1031 \tabularnewline
IOU+Motion   & 36.2  & 37.7  & 52  & 253  & \textbf{5707}  & 65056  & 913 \tabularnewline
IOU+Motion+Deep\cite{Yu2016POI}  & 39.9  & 43.8  & 72  & \textbf{220}  & 7209  & 59649  & 645 \tabularnewline
\hline 
Explicit features (Eq. \ref{eq:cost_function})  & 40.1  & 47.1  & 73  & 226  & 7133  & \textbf{59623}  & 496 \tabularnewline
Implicit features (Eq. \ref{eq: GRU}) & \textbf{40.2}  & \textbf{49.6}  & \textbf{74}  & 231  & 6462  & 59706  & \textbf{380} \tabularnewline
\hline 
\end{tabular}
\end{table*}
\begin{table*}[tbh]
\caption{Analysis on the online MOT16 validation benchmark.\label{tab:MOT16_result} }
\centering{} %
\begin{tabular}{|c|c|c|c|c|c|c|c|c|}
\hline 
Trackers  & MOTA$\uparrow$  & MT$\uparrow$  & ML$\downarrow$  & FP$\downarrow$  & FN$\downarrow$  & IDs$\downarrow$  & Frag$\downarrow$  & Hz $\uparrow$ \tabularnewline
\hline 
RAR16pub\cite{Kuan2017Recurrent}  & 45.9  & 13.2\%  & 41.9\%  & 6871  & 91173  & 648  & 1992  & 0.9\tabularnewline
STAM16 \cite{Chu2017Online}  & 46.0  & 14.6\%  & 43.6\%  & 6895  & 91117  & \textbf{473}  & \textbf{1422}  & 0.2 \tabularnewline
DMMOT \cite{Zhu2018Online}  & 46.1  & \textbf{17.4}\%  & 42.7\%  & 7909  & 89874  & 532  & 1616  & 0.3 \tabularnewline
AMIR\cite{Sadeghian2017Tracking}  & 47.2  & 14.0\%  & 41.6\%  & \textbf{2681}  & 92856  & 774  & 1675  & 1 \tabularnewline
\hline 
GAKP  & \textbf{48.1}  & 14.1\%  & \textbf{38.2}\%  & 7413  & \textbf{85971}  & 729  & 1964  & \textbf{7.8}\tabularnewline
\hline 
\end{tabular}
\end{table*}

\begin{table*}[!h]
\caption{Analysis on the online MOT17 validation benchmark.\label{tab:MOT_result} }
\centering{} %
\begin{tabular}{|c|c|c|c|c|c|c|c|c|c|}
\hline
Trackers  & MOTA$\uparrow$  & ID F1$\uparrow$  & MT$\uparrow$  & ML$\downarrow$  & FP$\downarrow$  & FN$\downarrow$  & IDs$\downarrow$  & Frag$\downarrow$  & Hz$\uparrow$ \tabularnewline
\hline 
yt\_face  & 52.6  & 51.5  & \textbf{23.0}\%  & 35.9\%  & 23,894  & \textbf{241,489}  & 2,047  & \textbf{2,827}  & 2.2 \tabularnewline
IOUT\_Re  & \textbf{52.7}  & 43.3  & 20.1\%  & \textbf{32.6}\%  & \textbf{16,529}  & 243,226  & 6,946  & 6,520  & 7.0 \tabularnewline
JCC  & 51.2  & \textbf{54.5}  & 20.9\%  & 37.0\%  & 25,937  & 247,822  & \textbf{1,802}  & 2,984  & 1.8 \tabularnewline
FWT  & 51.3  & 47.6  & 21.4\%  & 35.2\%  & 24,101  & 247,921  & 2,648  & 4,279  & 0.2 \tabularnewline
MHT\_DAM  & 50.7  & 47.2  & 20.8\%  & 36.9\%  & 22,875  & 252,889  & 2,314  & 2,865  & 0.9 \tabularnewline
EDMT17  & 50.0  & 51.3  & 21.6\%  & 36.3\%  & 32,279  & 247,297  & 2,264  & 3,260  & 0.6 \tabularnewline
\hline
GAKP  & 51.6  & 53.9  & 17.3\%  & 35.5\%  & 21,419  & 249,059  & 2,384  & 5,613  & \textbf{7.8} \tabularnewline
\hline 
\end{tabular}
\end{table*}

\subsection{Ablation Study \label{ablation}}


The underlying motivation of our proposed framework is to address the challenge: optimal combination of multiple features, and the disadvantage of few training data in deep learning. We now present experiments towards two goals on MOT benchmark.

\textbf{Combination of multiple features.} 
One advantage of our work compared with prior works is optimal combination of features. We investigate different combination of features in our tracking framework by measuring the performance in terms of MOTA on the MOT$17$ training sequences.
The auto-tuning Kalman \cite{AutoKalman2018} prediction in Section \ref{subsec:Motion-Model} is applied in all experiments. 
IOU-based data association mentioned in \cite{Bochinski2017High} are used as our baseline. And IOU and motion combined data association is formulated as our contrast experiment with motion and spatial model, respectively. Finally we explicitly combines different features proposed in \cite{Yu2016POI}. 

The comparison results are shown in Fig. \ref{fig:compare}. 
The motion information helps to increase performance by $1.1\%$ over baseline in the contrast experiment. 
We can observe that the method in \cite{Yu2016POI} outperforms Kalman baseline by $4.8\%$ in terms of MOTA on MOT$17$-DPM training data set, which demonstrates appearance model based on deep feature is more powerful than traditional motion model.
The overall result is shown in Table \ref{tab:MOT_result-self}.

Our two proposed models with explicit/implicit features in this work show better performance than two baselines. Moreover, the implicit features calculated with pre-trained GRU achieve excellent performance over explicit feature model, where increase by 0.6\% in MOTA. The main reason is that the weights of feature are end-to-end learned from training data by GRU in the implicit feature model. In our implicit feature model, initial sub-image pairs are used to train the GRU network instead of hand-craft features, such as Euclidean distance, IOU, deep feature and so on. 

\textbf{Impact of few training data}. 
One of the advantages of our representation compared with the previous is the capacity to compensate the lack of training data.  We investigate the performance compared with the online trackers AMIR \cite{Sadeghian2017Tracking}, RAR16pub  \cite{Kuan2017Recurrent}, STAM16 \cite{Chu2017Online} and DMMOT \cite{Zhu2018Online} in MOT$16$ validation benchmark. 
Table \ref{tab:MOT16_result} shows the details of these published online tracker for the validation sets. 
Our method achieves a competitive score and performs favorably against to others. Our proposed method outperforms the $RNN\_LSTM$ based tracker (AMIR) by $0.9\%$ in MOTA, $0.1\%$ in MT, $3.4\%$ in ML and $7.4\%$ in FN, respectively. The main reason is the proposed method utilize numerous online generated training data, while the issue of the small size of MOT$16$ training data is not well considered in AMIR.  

The spatial-temporal attention network utilized in the trackers DMMOT$\&$STAM16 is trained on MOT$15\&16$ datasets. The recurrent autoregressive network parameters of tracker RAR16pub are learned from discriminate ground truth associations and false associations in MOT data.
The MOTA score drops significantly by $2.0\%$ compared with our method, as tracking videos have only $1221$ and $1276$ object trajectories in MOT$15$ and MOT$16$ respectively. 


\subsection{Comparison with State-of-the-art Algorithms\label{result}}

In order to further validate our proposed algorithm, we compare the state-of-the-art methods on MOT$17$ benchmark, and the results are presented in Table  \ref{tab:MOT_result}. Obviously, our method achieves competitive performance of the comprehensive evaluation metric:  MOTA=$51.6$, which ranks 3rd amongst all the online MOT approaches. 

We notice that MT is an evaluation metric for which the proposed method performs worst when compared with other state-of-the-art methods. MT is dependent on the detection results given by the MOT benchmark which are noisy with the false positive and false negative. We believe a better pedestrian filter instead of the public detector used in the benchmark will help to improve the MT metric. Beside, it is evident that our proposed GAKP algorithm outperforms all other top competitors in terms of efficiency, \ie, $Hz=7.8$. 

\section{Conclusion}

In this work, we proposed a novel tracking method that integrates the GRU and the auto-tuning Kalman for MOT, achieving a near-optimum with only a small amount of training data. Experimental results show that our algorithm can achieve competitive performance on the challenging MOT benchmark with faster process speed compared to the state-of-the-art RNN-based online MOT algorithms.
{\small
\bibliographystyle{named}
\bibliography{ijcai21}
}

\end{document}